\documentclass[11pt,a4paper]{article}
\usepackage{times,latexsym}
\usepackage{url}
\usepackage[T1]{fontenc}

\usepackage[acceptedWithA]{tacl2018v2}

\usepackage{xspace,mfirstuc,tabulary}

\newif\iftaclinstructions
\taclinstructionsfalse %
\iftaclinstructions
\renewcommand{\confidential}{}
\renewcommand{\anonsubtext}{(No author info supplied here, for consistency with
TACL-submission anonymization requirements)}
\newcommand{\instr}
\fi

\usepackage{times}
\usepackage{latexsym}
\usepackage{amsmath}
\usepackage[T1]{fontenc}

\usepackage[utf8]{inputenc}

\usepackage{microtype}

\usepackage{soul} %

\usepackage{graphicx} %
\usepackage{subfig} %
\usepackage{booktabs} %
\usepackage{multirow} %
\usepackage{comment} %
\usepackage{amssymb} %
\usepackage{multicol}
\usepackage{dsfont} %

\usepackage[shortlabels]{enumitem} %

\iftaclpubformat %

\else

\fi

\definecolor{WildStrawberry}{HTML}{ffab91}
\definecolor{OrangeRed}{HTML}{ff7043}

\definecolor{LightBlue}{HTML}{98F5FF}

\DeclareRobustCommand{\hltrue}[1]{{\sethlcolor{LightBlue}\hl{#1}}}
\DeclareRobustCommand{\hlfalseo}[1]{{\sethlcolor{pink}\hl{#1}}}
\DeclareRobustCommand{\hlfalset}[1]{{\sethlcolor{WildStrawberry}\hl{#1}}}
\DeclareRobustCommand{\hlfalsetr}[1]{{\sethlcolor{OrangeRed}\hl{#1}}}

\newcommand{\resource}{\textsc{ParaRel}\raisebox{-2pt}{\includegraphics[width=0.15in]{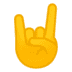}}}

\title{Measuring and Improving Consistency in Pretrained Language Models}

 \author{Yanai Elazar\textsuperscript{1,2} \,
 Nora Kassner\textsuperscript{3} \,
 Shauli Ravfogel\textsuperscript{1,2} \,
 Abhilasha Ravichander\textsuperscript{4} \, \\
 {\bf Eduard Hovy\textsuperscript{4}\,
 \bf Hinrich Sch\"utze\textsuperscript{3}\,
 Yoav Goldberg\textsuperscript{1,2}}\\
\textsuperscript{1}Computer Science Department, Bar Ilan University \\
\textsuperscript{2}Allen Institute for Artificial Intelligence \\
\textsuperscript{3}Center for Information and Language Processing (CIS), LMU Munich\\
\textsuperscript{4}Language Technologies Institute, Carnegie Mellon University \\
  {\tt  \{yanaiela,shauli.ravfogel,yoav.goldberg\}@gmail.com}\\
  {\tt kassner@cis.lmu.de}
  {\tt \{aravicha,hovy\}@cs.cmu.edu}
  }

\date{}

\newcounter{notecounter}

\newcommand{\enoteson}{\long\gdef\enote##1##2{{
\stepcounter{notecounter}
{\large\bf
\hspace{1cm}\arabic{notecounter} $<<<$ ##1: ##2
$>>>$\hspace{1cm}}}}}
\enoteson

\begin{document}
\maketitle

\begin{abstract}

\textit{Consistency} of a model --- that is, the invariance
of its behavior under meaning-preserving alternations in its
input --- is a highly desirable property in natural language
processing.  In this paper we study the question: Are
Pretrained Language Models (PLMs) consistent with respect to
factual knowledge?
To this end, we create \resource{}, a
high-quality resource of cloze-style query English
paraphrases. It contains a total of 328 paraphrases for 38 relations. Using \resource{}, we show that the consistency
of all PLMs we experiment with is poor -- though with high
variance between relations.  Our analysis of the
representational spaces of PLMs suggests that they have a
poor structure and are currently not suitable for
representing knowledge robustly. Finally, we propose
a method for improving model consistency and experimentally
demonstrate its effectiveness.\footnote{The code and resource are available at: \url{https://github.com/yanaiela/pararel}.}

\end{abstract}

\section{Introduction}
\label{sec:intro}

Pretrained Language Models (PLMs) are
large neural networks that are
used in a wide variety of NLP tasks. They operate under a
pretrain-finetune paradigm: models are first \emph{pretrained} over a large text corpus and then \emph{finetuned} on a downstream task. PLMs are thought of as good language encoders, supplying basic language understanding capabilities that can be used with ease for many downstream tasks.

A desirable property of a good language understanding model
is \emph{consistency}: the ability to make consistent
decisions in semantically equivalent contexts, reflecting a
systematic ability to generalize in the face of language variability.

Examples of consistency include: predicting the same answer in question answering and reading comprehension tasks regardless of paraphrase \cite{consistent-qa}; making consistent assignments in coreference resolution \cite{denis2009global,chang2011inference}; or making summaries factually consistent with the original document \cite{kryscinski2020evaluating}.
While consistency is important in many tasks, nothing in the training process explicitly targets it. One could hope that
the unsupervised training signal from large corpora
made available to PLMs such as BERT or RoBERTa
\cite{bert,roberta} is sufficient to induce consistency and
transfer it to downstream tasks.
In this paper, we show that this is not the case.

\begin{figure}[t!]
\centering

\includegraphics[width=1.\columnwidth]{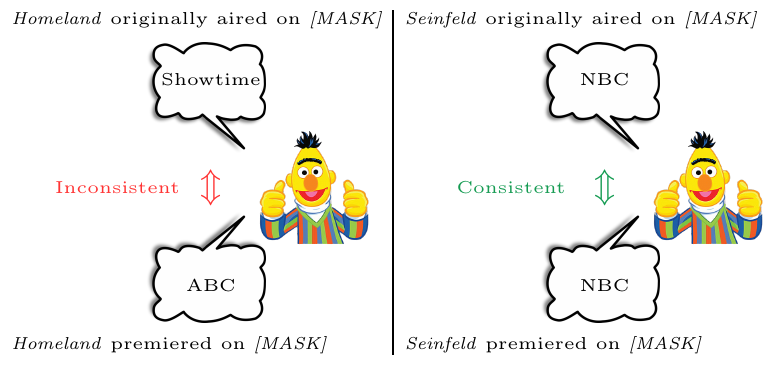}

\caption{Overview of our approach. 
We expect that a consistent model would predict the same answer for %
 two paraphrases.
In this example, the model is inconsistent on the
\textit{Homeland} and consistent on the \textit{Seinfeld} paraphrases.}
\label{fig:overview}
\end{figure}

The recent rise of PLMs has sparked a discussion about whether these models can be used as Knowledge Bases (KBs) \cite{lama,petroni2020how,davison2019commonsense,peters2019knowledge,alpaqa,roberts2020much}. 
Consistency is a key property of KBs and is particularly important for automatically constructed KBs. %
One of the biggest appeals of using a PLM as a KB is that we can
query it in natural language -- instead of relying on a specific KB schema.
The expectation is that PLMs abstract away from language and map queries in natural language into meaningful representations such that queries with identical intent but different language forms yield the same answer. 
For example, the query ``\textit{Homeland} premiered on \textit{[MASK]}'' should produce the same answer as ``\textit{Homeland} originally aired on \textit{[MASK]}''.
Studying inconsistencies of PLM-KBs can also teach us about the organization of knowledge in the model or lack thereof. 
Finally, failure to behave consistently may point
to other representational issues such as the similarity between antonyms and synonyms \cite{nguyen2016integrating}, and overestimating events and actions (reporting bias) \cite{shwartz2020neural}.

In this work, we study the consistency of factual knowledge
in PLMs, specifically in Masked Language Models (MLM) -- these are PLMs trained with the MLM objective \cite{bert,roberta}, as opposed to other strategies such as standard language modeling \cite{gpt2} or text-to-text \cite{t5}. We ask: Is the factual information we extract from PLMs invariant to paraphrasing? We use zero-shot evaluation since we want to inspect models directly, without adding biases through finetuning. This allows us to assess how much consistency was acquired during pretraining and to compare the consistency of different models. Overall, we find that the consistency of the PLMs we consider is poor, although there is a high variance between relations.

We introduce \resource{}, a new benchmark that enables us to measure consistency in PLMs (\S \ref{sec:probe}), by using factual knowledge that was found to be partially encoded in them \cite{lama,alpaqa}.
\resource{} is a manually curated resource
that provides patterns -- short textual prompts -- that are paraphrases of one another, with 328 paraphrases describing 38 binary relations such as \textit{X born-in Y}, \textit{X works-for Y} (\S \ref{sec:rel-graph}).
We then test multiple PLMs for knowledge consistency, i.e., whether
a model predicts the same answer for all patterns of a relation.
Figure \ref{fig:overview} shows an overview of our approach.
Using \resource{}, we probe for consistency in four PLM
types: BERT, BERT-whole-word-masking, RoBERTa and ALBERT (\S
\ref{sec:setup}).
Our experiments with \resource{} show that
current models have poor consistency, although with high variance between relations (\S \ref{sec:experiments}). 

Finally, we propose a method that improves model consistency
by introducing a novel consistency loss
(\S \ref{sec:adding_consistency}). We demonstrate that
trained with this loss, BERT achieves better consistency performance on unseen relations. However, more work is required to achieve fully consistent models.

\section{Background}
\label{sec:background}
There has been significant interest in analyzing how well
PLMs \cite{rogers2020primer} perform on
linguistic tasks
\cite{yoav-syntax,hewitt2019structural,tenney2019bert,amnesic_probing},
commonsense \cite{forbes2019neural,
  da2019cracking,zhang2020language} and reasoning
\cite{talmor2019olmpics,
  kassner-etal-2020-pretrained}, usually assessed by
measures of accuracy.
However, accuracy is just one measure of PLM performance \cite{linzen2020can}. It is
equally important that PLMs do not make contradictory
predictions (cf.\ Figure \ref{fig:overview}), a type of error that humans rarely make. 
There has been relatively little research attention devoted to this question, i.e., to analyze if models behave \emph{consistently}.
One example concerns negation:
\citet{Ettinger_2020} and \citet{kassner-schutze-2020-negated}
show
that models tend to generate facts and their negation, a
type of inconsistent behavior.
\newcite{ravichander-etal-2020-systematicity}
propose paired probes for evaluating consistency.
 Our work is
broader in scope, examining the consistency of PLM behavior across a
range of factual knowledge types and investigating how
models can be made to behave more consistently.

Consistency has also been highlighted as a desirable
property in automatically constructed KBs and downstream NLP
tasks. We now briefly review work along these lines.

\textbf{Consistency in knowledge bases (KBs)} has been
studied in theoretical frameworks in the context of the
satisfiability problem and KB construction, and efficient
algorithms for detecting inconsistencies in KBs have been
proposed \cite{hansen2000probabilistic,andersen2001easy}.
Other work aims to quantify the degree to which KBs are
inconsistent and detects inconsistent statements
\cite{Thimm:2009d,Thimm:2013,muino2011measuring}.

\textbf{Consistency in question answering} was studied by
\citet{ribeiro-etal-2019-red} in two tasks: visual question answering \cite{vqa} and reading comprehension \cite{squad}. They automatically generate questions to test the consistency of QA models.
Their findings suggest that most models are not consistent in their predictions. In addition, they use data augmentation to create more robust models.
\citet{alberti2019synthetic} generate new questions
conditioned on context and answer from a labeled dataset
and by filtering answers that do not provide a consistent
result with the original answer. They show that pretraining on these synthetic data improves QA results.
\citet{consistent-qa}  use data augmentation that complements questions with symmetricity and transitivity, as well as a regularizing loss that penalizes inconsistent predictions.
\citet{DBLP:journals/corr/abs-2104-08401} propose a method to improve accuracy and consistency of QA models by augmenting a PLM with an evolving memory that records PLM answers and resolves inconsistency between answers.

Work on \textbf{consistency in other domains}
includes \citep{du2019consistent} where  prediction of
consistency in procedural text is improved. \citet{ribeiro-etal-2020-beyond} use consistency for more robust evaluation. \citet{li-etal-2019-logic} measure and mitigate inconsistency in natural language inference (NLI), and finally, \citet{camburu2020make} propose a method for measuring inconsistencies in NLI explanations \cite{camburu2018snli}.

\section{Probing PLMs for Consistency}
\label{sec:probe}

In this section, we formally define consistency and describe our framework for probing consistency of PLMs.

\subsection{Consistency}
We define a model as \emph{consistent} if, given  two
\textit{cloze-phrases} such as 
 ``\textit{Seinfeld} originally aired on \textit{[MASK]}'' and
``\textit{Seinfeld} premiered on \textit{[MASK]}'' that
are \textit{quasi-paraphrases}, it makes non-contradictory
predictions\footnote{We refer to \textit{non-contradictory
    predictions} as predictions that, as the name suggest,
  do not contradict one another. For instance, predicting as
  the birth place of a person two difference cities is
  considered to be contradictory, but predicting a city and
  its country, is \textbf{not}.}
on N-1 relations over a large set of entities.
A \textit{quasi-paraphrase} -- a concept introduced by \citet{what_is_paraphrase} -- is a more fuzzy version of a paraphrase. The concept does not rely on the strict, logical definition of paraphrase and allows to operationalize concrete uses of paraphrases. This definition is in the spirit of the RTE definition \cite{dagan-rte}, which similarly supports a more flexible use of the notion of entailment.
For instance, a model that predicts \textit{NBC} and \textit{ABC} on the two aforementioned patterns, is not consistent, since these two facts are contradictory. We define a \textit{cloze-pattern} as a cloze-phrase that expresses a relation between a subject and an object.
Note that consistency does not require the answers to be factually correct. While correctness is also an important property for KBs, we view it as a separate objective and measure it independently.
We use the terms \textit{paraphrase} and \textit{quasi-paraphrase} interchangeably.

Many-to-many (N-M) relations (e.g. \textit{shares-border-with}) can be consistent 
even with different answers (given they are correct). For
instance, two patterns that express the
\textit{shares-border-with} relation and predict
\textit{Albania} and \textit{Bulgaria} for
\textit{Greece}
are both correct. We do not consider such relations for measuring consistency. However, another requirement from a KB is \textit{determinism}, i.e., returning the results in the same order (when more than a single result exists).
In this work, we focus on consistency, but also measure determinism of the models we inspect.

\subsection{The Framework}
\label{sec:framework}

\begin{figure*}[t!]
\centering

\includegraphics[width=1.\textwidth]{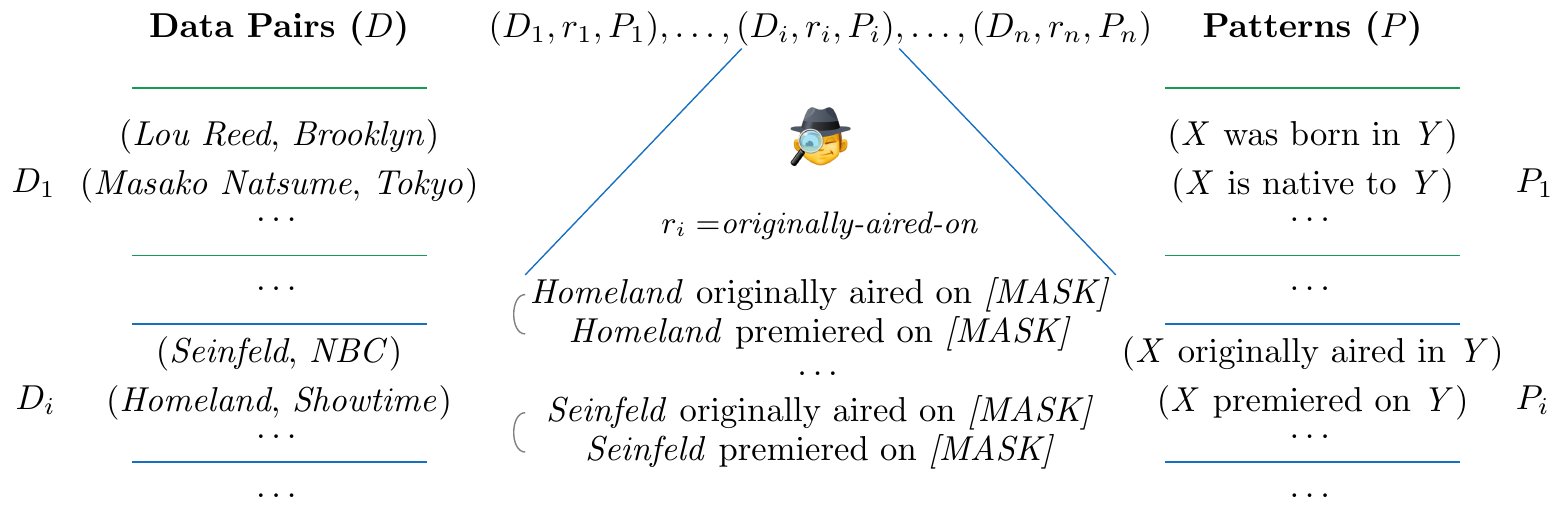}

\caption{Overview of our framework for assessing model
  consistency. $D_i$ (``Data Pairs $(D)$'' on the left) is a
  set of KB triplets of some relation $r_i$, which are
  coupled with a set of \textit{quasi-paraphrase}
  cloze-patterns $P_i$
(``Patterns $(P)$'' on the right)
  that describe that relation. We then populate the subjects
  from $D_i$ as well as a mask token into all patterns $P_i$
(shown in the middle)
  and expect a model to predict the same object across all pattern pairs.}
\label{fig:framework}
\end{figure*}

An illustration of the framework is presented in Figure \ref{fig:framework}.
Let $D_i$ be a set of subject-object KB tuples (e.g., <\textit{Homeland}, \textit{Showtime}>) from some relation $r_i$ (e.g., \textit{originally-aired-on}), accompanied with a set of \textit{quasi-paraphrases} cloze-patterns $P_i$ (e.g., \textit{X} originally aired on \textit{Y}).
Our goal is to test whether the model consistently predicts the same object (e.g., \textit{Showtime}) for a particular subject (e.g., \textit{Homeland}).\footnote{Although it is possible to also predict the subject from the object, in the cases of N-1 relations more than a single answer would be possible, making it impossible to test for consistency, but determinism.} To this end, we substitute \textit{X} with a subject from $D_i$ and \textit{Y} with \textit{[MASK]} in all of the patterns $P_i$ of that relation (e.g., \textit{Homeland} originally aired on \textit{[MASK]} and \textit{Homeland} premiered on \textit{[MASK]}).
A consistent model must predict the same entity.

\paragraph{Restricted Candidate Sets}

Since PLMs were not trained for serving as KBs, they often predict words that are not KB entities; e.g., a PLM may predict, for the pattern ``\textit{Showtime} originally aired on \textit{[MASK]}'', the noun `tv' --  which is also a likely substitution for the language modeling objective, but not a valid KB fact completion.
Therefore, following \citep{Xiong2020Pretrained, ravichander-etal-2020-systematicity, kassner2021multilingual}, we restrict the PLMs' output vocabulary to the set of possible gold objects for each relation from the underlining KB. For example, in the \textit{born-in} relation, instead of inspecting the entire vocabulary of a model, we only keep objects from the KB, such as \textit{Paris}, \textit{London}, \textit{Tokyo}, etc.

Note that this setup makes the task easier for the PLM,
especially in the context of KBs. However, poor
consistency in this setup strongly implies that consistency
would be even lower without restricting candidates.

\section{The \resource{} Resource}
\label{sec:rel-graph}

We now describe \resource{}, a resource designed for our framework (cf.\ Section \ref{sec:framework}).
\resource{} is curated by experts, with a high level of agreement.
It contains patterns for 38 relations\footnote{using the 41 relations from LAMA \cite{lama}, leaving out three relations that are poorly defined, or consist of mixed and heterogeneous entities.} from T-REx \cite{trex} --- a large dataset containing KB triples aligned with Wikipedia abstracts --- with an average of 8.63 patterns per relation.
Table \ref{tab:rel-graph-stats} gives statistics.
We further analyse the paraphrases used in this resource, partly based on the types defined in \citet{what_is_paraphrase}, and report this analysis in Appendix \ref{sec:paraphrase_analysis}.

\paragraph{Construction Method}
\resource{} was constructed in four steps. (1) We began with
the patterns provided by LAMA \cite{lama} (one pattern per
relation, referred to as \textit{base-pattern}). (2) We augmented each base-pattern with other patterns that are paraphrases from
LPAQA \cite{alpaqa}. However, since LPAQA was
created automatically (either by back-translation or by extracting patterns from sentences that contain both subject and object), some LPAQA patterns are not
correct paraphrases.
We therefore only include the subset of correct paraphrases.
(3) Using SPIKE
\cite{spike},\footnote{\url{https://spike.apps.allenai.org/}}
a search engine over Wikipedia sentences that supports
syntax-based queries, we searched for additional patterns
that appeared in Wikipedia and added them to 
\resource{}. Specifically, we searched for Wikipedia sentences
containing a  subject-object
tuple from T-REx and then manually extracted 
patterns from the sentences. (4) Lastly, we added
additional paraphrases of the base-pattern
using the annotators' linguistic expertise. Two additional
experts went over all the patterns and corrected them, while
engaging in a discussion until reaching agreement,
discarding patterns they could not agree on.

\paragraph{Human Agreement}
To assess the quality of \resource{}, we run a
human annotation study.
For each relation, we sample up to
five paraphrases, comparing each of the new patterns to the
base-pattern from LAMA.
That is, if relation $r_i$ contains the following patterns:
$p_1, p_2, p_3, p_4$, and $p_1$ is the
base-pattern, then we compare the following pairs $(p_1, p_2), (p_1, p_3), (p_1,p_4)$.

We populate the patterns with random subjects and objects pairs from T-REx \cite{trex} and ask annotators if these sentences are paraphrases.
We also sample patterns from different relations to provide
examples that are not paraphrases of each other, as a control.
Each task contains five patterns that are thought to be paraphrases and two that are not.\footnote{The controls contain the same subjects and objects so that only the pattern (not its arguments) can be used to solve the task.}
Overall, we collect annotations for 156 paraphrase candidates and 61 controls.

We asked NLP graduate students to annotate the pairs and collected one answer per pair.\footnote{We asked the annotators to re-annotate any mismatch with our initial label, to allow them to fix random mistakes.}
The agreement scores for the paraphrases and the controls
are 95.5\% and 98.3\%, which is high and
indicates \resource's high quality.
We also inspected  the disagreements  %
and fixed many additional problems %
to further improve quality.

\begin{table}[t]
    \centering
\begin{tabular}{lr}
  \toprule
    \# Relations &   38 \\
    \# Patterns & 328 \\
    \midrule
 Min \# patterns per rel. &    2 \\
 Max \# patterns per rel. &   20 \\
 Avg \# patterns per rel. &    8.63 \\
 \midrule
   Avg syntax  &    4.74 \\
  Avg lexical  &    6.03 \\
\bottomrule
\end{tabular}
    \caption{Statistics of \resource{}. 
    Last two rows: average number of unique
    syntactic/lexical variations of patterns for a relation.}
    \label{tab:rel-graph-stats}
    
    \vspace{-0.1in}
\end{table}

\section{Experimental Setup}
\label{sec:setup}

\subsection{Models \& Data}
\label{setupdata}
We experiment with four PLMs:
BERT, BERT whole-word-masking\footnote{BERT whole-word-masking is BERT's version where words that are tokenized into multiple tokens are masked together.}
\cite{bert}, RoBERTa \cite{roberta} and ALBERT
\cite{albert}. For BERT, RoBERTa and ALBERT, we use a base and a large version.\footnote{For ALBERT we use the smallest and largest versions.}
We also report a majority baseline that always predicts the most common object for a relation. By construction, this baseline is perfectly consistent.

We use knowledge graph data from T-REx \cite{trex}.\footnote{We discard three poorly defined relations from T-REx.} To make the results comparable across models, we remove objects that are not represented as a single token in all models'
vocabularies; 26,813 tuples remain.\footnote{In a few cases, we filter entities from certain relations that contain multiple fine-grained relations to make our patterns compatible with the data. For instance, most of the instances for the \textit{genre} relation describes music genres, thus we remove some of the tuples were the objects include non-music genres such as `satire', `sitcom' and `thriller'.}
We further split the data into N-M relations for which we
report  determinism results (seven relations) and N-1
relations for which we report consistency (31 relations).

\subsection{Evaluation}
\label{sec:eval}

Our consistency measure for a relation $r_i$ (\textit{Consistency}) is the
percentage of consistent predictions of all the pattern pairs
$p_k^i,p_l^i \in P_i$ of that relation, for all its KB tuples
$d^i_j \in D_i$. Thus, for each KB tuple from a relation
$r_i$ that contains $n$ patterns, we consider predictions
for $n(n-1)/2$ pairs.

We also report \textit{Accuracy}, that is, the acc@1 of a model in predicting the correct object, using the original patterns from \citet{lama}. In contrast to \citet{lama}, we define it as the accuracy of the top-ranked object from the candidate set of each relation.
Finally, we report \emph{Consistent-Acc}, a new measure that evaluates individual objects as correct only
if \emph{all} patterns of the corresponding relation predict the object
correctly. \textit{Consistent-Acc} is  much stricter  and combines the
requirements of both consistency (\textit{Consistency}) and factual correctness (\textit{Accuracy}).
We report the average over relations, i.e., macro average, but notice that the micro average produces similar results.

\section{Experiments and Results}
\label{sec:experiments}

\begin{table}[t]
    \centering
\resizebox{1\columnwidth}{!}{%
\begin{tabular}{lrrrr}
\toprule
Model &    Succ-Patt &   Succ-Objs &  Unk-Const &   Know-Const \\

\midrule
majority       &   97.3+-7.3 &  23.2+-21.0 &  100.0+-0.0 &  100.0+-0.0 \\
\midrule
BERT-base      &  \textbf{100.0}+-0.0 &  63.0+-19.9 &  46.5+-21.7 &  63.8+-24.5 \\
BERT-large     &  \textbf{100.0}+-0.0 &  \textbf{65.7}+-22.1 &  48.1+-20.2 &  65.2+-23.8 \\
BERT-large-wwm &  \textbf{100.0}+-0.0 &  64.9+-20.3 &  \textbf{49.5}+-20.1 &  \textbf{65.3}+-25.1 \\
\midrule
RoBERTa-base   &  \textbf{100.0}+-0.0 &  56.2+-22.7 &  43.9+-15.8 &  56.3+-19.0 \\
RoBERTa-large  &  \textbf{100.0}+-0.0 &  60.1+-22.3 &  46.8+-18.0 &  60.5+-21.1 \\
\midrule
ALBERT-base    &  \textbf{100.0}+-0.0 &  45.8+-23.7 &  41.4+-17.3 &  56.3+-22.0 \\
ALBERT-xxlarge &  \textbf{100.0}+-0.0 &  58.8+-23.8 &  40.5+-16.4 &  57.5+-23.8 \\
\bottomrule
\end{tabular}

}
    \caption{Extractability Measures in the different models we inspect. Best model for each measure  highlighted in bold.}
    \label{tab:extractability_results}
\end{table}

\subsection{Knowledge Extraction through Different Patterns}

We begin by assessing our patterns as
well as the degree to which they extract the correct
entities. These results are summarized in Table
\ref{tab:extractability_results}.

First, we report \emph{Succ-Patt}, the percentage of
patterns that successfully predicted the right object at
least once. A high score suggests that the patterns are of high quality and enable the models to extract the correct answers. All PLMs achieve a perfect score.
Next, we report
\emph{Succ-Objs},
the percentage of entities that were predicted correctly by at least one of the patterns.
\textit{Succ-Objs} quantifies the degree to which the models
``have'' the required knowledge.
We observe that some tuples are not predicted correctly by any of our patterns: the scores vary between 45.8\% for ALBERT-base and 65.7\% for BERT-large. %
With an average number of 8.63 patterns per relation, there are multiple ways to extract the knowledge, we thus interpret these results as evidence that a large part of T-REx knowledge is not stored in these models.

Finally, we measure \emph{Unk-Const}, a consistency measure for the subset of tuples
for which no pattern predicted the correct answer; %
and \emph{Know-Const}, consistency for the subset where
at least one of the patterns for a specific
relation predicted the correct answer.
This split into subsets is
based on \textit{Succ-Objs}.
Overall, the results indicate that when the factual knowledge is successfully extracted, the model is also more consistent.
For instance, for BERT-large, \textit{Know-Const}  is 65.2\% and \textit{Unk-Const} is 48.1\%.

\subsection{Consistency \& Knowledge}
\begin{table}[t]
    \centering
\resizebox{1\columnwidth}{!}{%
\begin{tabular}{lrrr}
\toprule
Model &        Accuracy & Consistency & Consistent-Acc \\
\midrule
majority       &  23.1+-21.0 &  100.0+-0.0 &  23.1+-21.0 \\
\midrule
BERT-base      &  45.8+-25.6 &  58.5+-24.2 &  27.0+-23.8 \\
BERT-large     &  48.1+-26.1 &  \textbf{61.1}+-23.0 &  \textbf{29.5}+-26.6 \\
BERT-large-wwm &  \textbf{48.7}+-25.0 &  60.9+-24.2 &  29.3+-26.9 \\
\midrule
RoBERTa-base   &  39.0+-22.8 &  52.1+-17.8 &  16.4+-16.4 \\
RoBERTa-large  &  43.2+-24.7 &  56.3+-20.4 &  22.5+-21.1 \\
\midrule
ALBERT-base    &  29.8+-22.8 &  49.8+-20.1 &  16.7+-20.3 \\
ALBERT-xxlarge &  41.7+-24.9 &  52.1+-22.4 &  23.8+-24.8 \\
\bottomrule
\end{tabular}

}
    \caption{Knowledge and Consistency Results. Best model for each measure  in bold.}
    \label{tab:consistency_results_small}
\end{table}

 In this section, we report
the overall knowledge measure that was used in \citet{lama} (\textit{Accuracy}),
the consistency measure (\textit{Consistency}),
and \textit{Consistent-\
  Acc}, which combines knowledge and consistency  (\textit{Consistent-Acc}). %
The results are summarized in Table
\ref{tab:consistency_results_small}.

We begin with the \textit{Accuracy} results. 
The results range between 29.8\% (ALBERT-base) and 48.7\% (BERT-large whole-word-masking).
Notice that our numbers differ from \citet{lama} as we use a
candidate set (\S\ref{sec:probe}) and only consider KB
triples whose object is a single token in all the PLMs we consider (\S\ref{setupdata}). 

Next, we report  \textit{Consistency}  (\S\ref{sec:eval}).
The BERT models achieve the highest scores. There is a consistent improvement from  {base} to {large} versions of each model.
In contrast to previous work that observed quantitative and qualitative improvements of RoBERTa-based models over BERT \cite{roberta,talmor2019olmpics}, in terms of consistency, BERT is more consistent than RoBERTa and ALBERT.
Still, the overall results are low (61.1\% for
the best model), %
even more remarkably so because the restricted candidate set
makes the task easier.
We note that the results are highly variant between models
 (performance on \textit{original-language} varies between
52\% and 90\%), and relations (BERT-large performance is
92\% on \textit{capital-of} and 44\% on \textit{owned-by}).

Finally, we report
\emph{Consistent-Acc}:
the results are much lower than for \textit{Accuracy}, as expected,
but follow similar trends: RoBERTa-base performs worse (16.4\%) and BERT-large best  (29.5\%).

Interestingly, we find strong correlations between Accuracy
and Consistency, ranging from 67.3\% for RoBERTa-base to
82.1\% for BERT-large (all with small p-values $\ll 0.01 $).

A striking result of the model comparison is
the clear superiority of
BERT, both in knowledge accuracy (which was also observed by \citet{autoprompt}) and knowledge
consistency. We hypothesize this result is caused by
  the different sources of training data: although Wikipedia
  is part of the training data for all models we consider,
  for BERT it is the main data source, but for RoBERTa and
  ALBERT it is only a small portion. Thus, when using
  additional data, some of the facts may be forgotten, or
  contradicted in the other corpora; this can diminish
  knowledge and compromise consistency behavior.
Thus, since Wikipedia is likely the largest unified source of factual knowledge that exists in unstructured data, giving it prominence in pretraining  makes it more likely that the model will incorporate Wikipedia's factual knowledge well.
These results may have a broader impact on models to
come: Training bigger models with more data (such as GPT-3 \cite{gpt3}) is not always beneficial.

\paragraph{Determinism}
We also measure determinism for N-M relations, i.e., we use
the same measure as \textit{Consistency}, but since difference predictions may be factually correct, these do not necessarily convey consistency violations, but indicate non-determinism. For brevity, we do not present all results, but the trend is similar to the consistency result (although not comparable, as the relations are different): 52.9\% and 44.6\% for BERT-large and RoBERTa-base, respectively.

\begin{table}[t]
    \centering
\resizebox{1\columnwidth}{!}{%
\begin{tabular}{lrrr}
\toprule
Model&        Acc & Consistency & Consistent-Acc \\

\midrule
majority               &  23.1+-21.0 &  100.0+-0.0 &  23.1+-21.0 \\
\midrule
RoBERTa-med-small-1M &   11.2+-9.4 &  37.1+-11.0 &    2.8+-4.0 \\
\midrule
RoBERTa-base-10M     &  17.3+-15.8 &  29.8+-12.7 &    3.2+-5.1 \\
RoBERTa-base-100M    &  22.1+-17.1 &  31.5+-13.0 &    3.7+-5.3 \\
RoBERTa-base-1B      &  \textbf{38.0}+-23.4 &  \textbf{50.6}+-19.8 &  \textbf{18.0}+-16.0 \\
\bottomrule
\end{tabular}
}
    \caption{Knowledge and consistency results for different RoBERTas, trained on increasing amounts of data. Best model for each measure in bold.}
    \label{tab:robertas}
\end{table}

\paragraph{Effect of Pretraining Corpus Size}
Next, we study the question of whether the number of tokens used during pretraining contributes to consistency.
We use the pretrained RoBERTa models from \citet{robertas} and repeat the experiments on four additional models.
These are RoBERTa-based models, trained on a sample of Wikipedia and the book corpus, with varying training sizes and parameters. We use one of the three published models for each configuration and report the average accuracy over the relations for each model in Table \ref{tab:robertas}.
Overall, \textit{Accuracy} and
\textit{Consistent-Acc} improve
with more training data.
However, there is an interesting outlier to this trend:
The model that was trained on one million tokens is more consistent than the models trained on ten and one-hundred million tokens. A potentially crucial difference is that this model has many fewer parameters than the rest (to avoid overfitting). It is nonetheless interesting that a model that is trained on significantly less data can achieve better consistency. On the other hand, it's accuracy scores are lower, arguably due to the model being exposed to less factual knowledge during pretraining.

\subsection{Do PLMs Generalize Over Syntactic Configurations?} 

Many papers have found neural models (especially PLMs) to naturally
encode syntax
\cite{linzen2016assessing,belinkov2017neural,marvin-linzen-2018-targeted,belinkov2019analysis,yoav-syntax,hewitt2019structural}.
Does this mean that PLMs have successfully abstracted
knowledge and can comprehend and produce it regardless of
syntactic variation?
We consider two scenarios. (1) Two patterns differ only in
syntax. (2) Both syntax and lexical choice are the same.
As a proxy, we define syntactic equivalence when the dependency path between the subject and object are identical.
We parse all patterns from \resource{} using a dependency parser \cite{spacy}\footnote{\url{https://spacy.io/}} and retain the path between the entities.
Success on (1) indicates that the model's knowledge processing is robust to syntactic variation. Success on (2) indicates that the model's knowledge processing is robust to
variation in word order and tense.

\begin{table}[t]
    \centering
\resizebox{0.8\columnwidth}{!}{%
\begin{tabular}{lll}
\toprule
Model & Diff-Syntax & No-Change \\
\midrule
majority       &            100.0+-0.0 &            100.0+-0.0 \\
\midrule
BERT-base      &            67.9+-30.3 &            76.3+-22.6 \\
BERT-large     &            67.5+-30.2 &            78.7+-14.7 \\
BERT-large-wwm &            63.0+-31.7 &             \textbf{81.1}+-9.7 \\
\midrule
RoBERTa-base   &            66.9+-10.1 &             80.7+-5.2 \\
RoBERTa-large  &            \textbf{69.7}+-19.2 &             80.3+-6.8 \\
\midrule
ALBERT-base    &            62.3+-22.8 &            72.6+-11.5 \\
ALBERT-xxlarge &            51.7+-26.0 &            67.3+-17.1 \\
\bottomrule
\end{tabular}

}
    \caption{Consistency and standard deviation when only syntax differs
      (\textit{Diff-Syntax}) and when syntax and lexical choice are identical (\textit{No-Change}). Best model for each metric is highlighted in bold.}
    \label{tab:syntax_results}
\end{table}

Table \ref{tab:syntax_results}
reports results.
While these and the main results on the entire dataset are not comparable as the pattern subsets are different,
they are higher than the general results: 67.5\% for BERT-large when only the syntax differs and 78.7\% when the syntax is
identical. This demonstrates that while PLMs have impressive syntactic abilities, they struggle to extract factual knowledge in the face of tense, word-order, and syntactic variation.

\citet{mccoy2019right}  show that supervised models trained on MNLI \cite{mnli}, an NLI
dataset \cite{snli}, use
superficial syntactic heuristics rather than more
generalizable properties of the data.
Our results indicate that PLMs have  problems along
the same lines:
they are not robust to surface variation.

\section{Analysis}
\label{sec:analysis}

\begin{table*}[t]
    \centering
\resizebox{1\textwidth}{!}{%
\begin{tabular}{lllllllll}
\toprule
\# & Subject & Object & Pattern \#1 & Pattern \#2 & Pattern \#3 & Pred \#1 & Pred \#2 & Pred \#3 \\
\midrule

1 & Adriaan Pauw & Amsterdam & [X] was born in [Y]. & [X] is native to [Y]. & [X] is a [Y]-born person. & \hltrue{Amsterdam} & \hlfalseo{Madagascar} & \hlfalset{Luxembourg} \\
2 & Nissan Livina Geniss & Nissan & [X] is produced by [Y]. & [X] is created by [Y]. & [X], created by [Y]. & \hltrue{Nissan} & \hlfalseo{Renault} & \hlfalseo{Renault} \\
3 & Albania & Serbia & [X] shares border with [Y]. & [Y] borders with [X]. & [Y] shares the border with [X] & \hlfalseo{Greece} & \hlfalset{Turkey} & \hlfalsetr{Kosovo} \\
4 & iCloud & Apple & [X] is developed by [Y]. & [X], created by [Y]. & [X] was created by [Y] & \hlfalseo{Microsoft} & \hlfalset{Google} & \hlfalsetr{Sony} \\
\midrule

5 & Yahoo! Messenger & Yahoo & [X], a product created by [Y] & [X], a product developed by [Y] & [Y], that developed [X] & \hlfalseo{Microsoft} & \hlfalseo{Microsoft} & \hlfalseo{Microsoft} \\
6 & Wales & Cardiff & The capital of [X] is [Y] . & [X]'s capital, [Y]. & [X]'s capital city, [Y]. & \hltrue{Cardiff} & \hltrue{Cardiff} & \hltrue{Cardiff} \\

\bottomrule
\end{tabular}

}
    \caption{Predictions of BERT-large-cased. ``Subject''
      and ``Object'' are  from T-REx \cite{trex}.
      ``Pattern \#$i$'' / ``Pred \#$i$'':
 three different patterns from our resource and
      their predictions. The predictions are colored in blue
      if the model predicted correctly (out of the candidate
      list), and in red otherwise. If there is more than a
      single erroneous prediction, it is colored by a different red.}
    \label{tab:predictions}
    
    \vspace{-0.1in}
\end{table*}

\subsection{Qualitative Analysis}
To better understand the factors affecting consistent
predictions, we inspect the predictions of BERT-large on the
patterns shown  in Table \ref{tab:predictions}.
We highlight several cases:
The predictions in Example \#1 are inconsistent, and correct for the first pattern (\textit{Amsterdam}), but not for the other two (\textit{Madagascar} and \textit{Luxembourg}). 
The predictions in Example \#2 also show a single pattern that predicted the right object; however, the two other patterns, which are lexically similar, predicted the same, wrong answer -- \textit{Renault}.
Next, the patterns of Example \#3
produced two factually correct answers out of three (\textit{Greece}, \textit{Kosovo}), but simply do not correspond to the gold
object in T-REx (\textit{Albania}), since this is an M-N relation. Note that this relation is not part of the consistency evaluation, but the determinism evaluation.
The three different predictions in example \#4 are all incorrect.
Finally, the two last predictions demonstrate consistent predictions:
Example \#5 is consistent but factually incorrect (even though the correct answer is a substring of the subject), and finally, Example \#6
is consistent and factual.

\begin{figure}[t!]
\centering
\includegraphics[width=1\columnwidth]{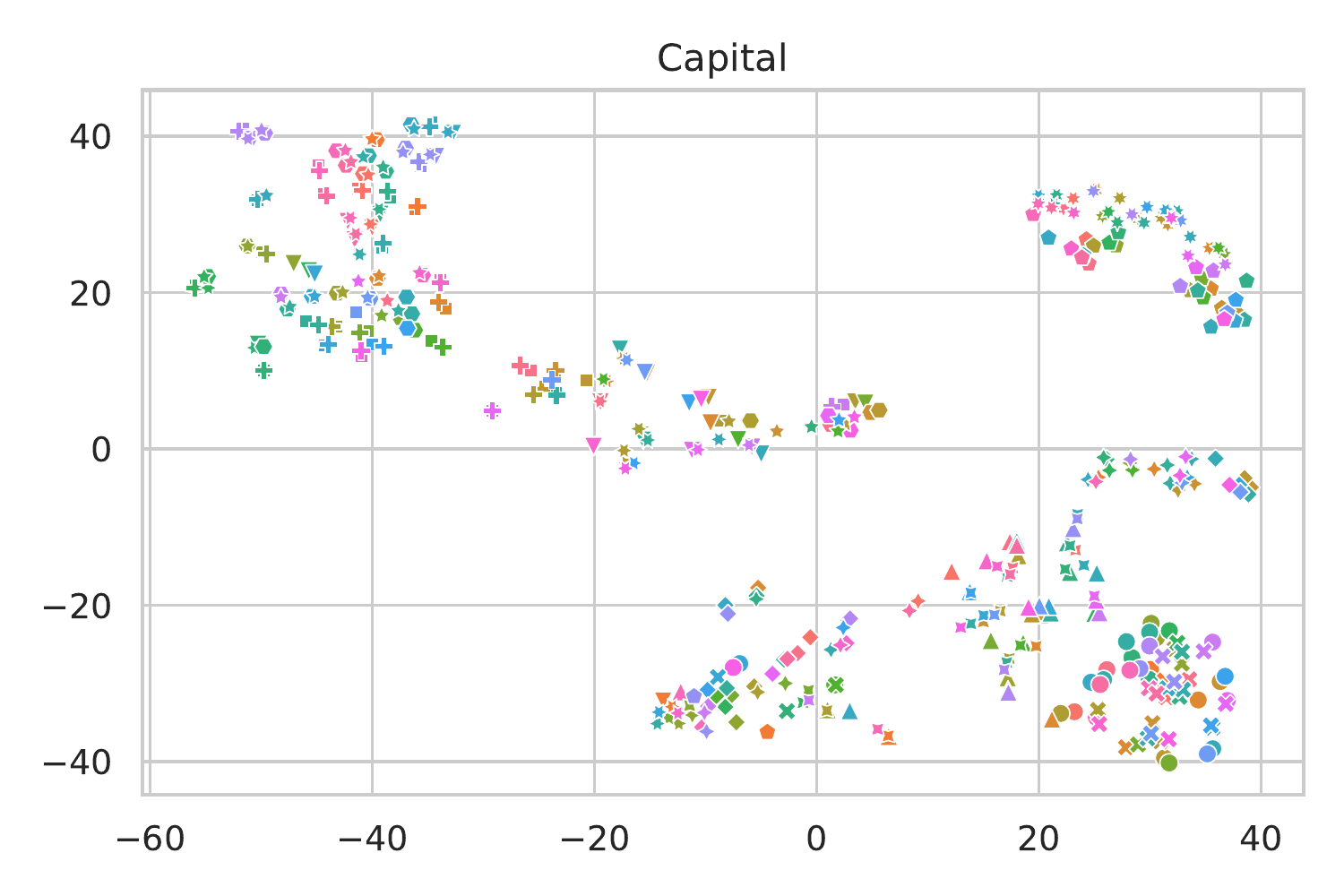}

\caption{t-SNE of the encoded patterns from the
  \textit{capital} relation. The colors
  represent the different subjects, while the shapes
  represent patterns. A knowledge-focused representation
  should cluster based on identical subjects (color), but
  instead the clustering is according to identical patterns (shape).}
\label{fig:tsne-emb}

\end{figure}

\subsection{Representation Analysis}

To provide insights on the models' representations, we inspect these after encoding the patterns.

Motivated by previous work that found that words with the same syntactic structure cluster together \cite{chi-etal-2020-finding,ravfogel-etal-2020-unsupervised} we perform a similar experiment to test if this behavior replicates with respect to knowledge:
We encode the patterns, after filling the placeholders with subjects and masked tokens and inspect the last layer representations in the masked token position.
When plotting the results using t-SNE \cite{tsne} we mainly
observe clustering based on the patterns, which suggests
that encoding of knowledge of the entity is not the main component of the representations.
Figure \ref{fig:tsne-emb} demonstrates
this for BERT-large encodings of the \textit{capital} relation, which is highly consistent.\footnote{While some patterns are clustered based on the subjects (upper-left part), most of them are clustered based on patterns.}
To provide a more quantitative assessment of this phenomenon, 
we also cluster the representations and set the number of centroids based on:\footnote{Using the KMeans algorithm.} (1) the number of patterns in each relation, which aims to capture pattern-based clusters, and (2) the number of subjects in each relation, which aims to capture entity-based clusters. This would allow for a perfect clustering, in the case of perfect alignment between the representation and the inspected property.
We measure the purity of these clusters using V-measure and observe that the clusters are mostly grouped by the patterns, rather than the subjects.
Finally, we compute the spearman correlation between the consistency scores and the V-measure of the representations.
However, the correlation between these variables is close to zero,\footnote{Except for BERT-large whole-word-masking, where the correlation is 39.5 ($p<0.05$).} therefore not explaining the models' behavior.
We repeated these experiments while inspecting the objects instead of the subjects, and found similar trends.
This finding is interesting since it means that (1) these representations are not knowledge-focused, i.e., their main component does not relate to knowledge, and (2) the representation by its entirety does not explain the behavior of the model, and thus only a subset of the representation does. %
This finding is consistent with previous work that observed similar trends for linguistic tasks \cite{amnesic_probing}.
We hypothesize that this disparity between the
representation and the behavior of the model may be explained by a situation where the distance between representations largely does not reflect the distance between predictions, but rather other, behaviorally irrelevant factors of a sentence.

\section{Improving Consistency in PLMs}
\label{sec:adding_consistency}

In the previous sections, we showed PLMs are generally not consistent in their predictions, and previous works have noticed the lack of this property in a variety of downstream tasks.
An ideal model would exhibit the consistency property after pretraining, and would then be able to transfer it to different downstream tasks. We therefore ask:
Can we enhance current PLMs and make them more consistent?

\subsection{Consistency Improved PLMs}
We propose to improve the consistency of PLMs by continuing the pretraining step with a novel consistency loss. %
We make use of the T-REx tuples and the paraphrases from \resource{}.

For each relation $r_i$, we
have a set of paraphrased patterns $P_i$ describing that relation.
We use a PLM to encode all patterns in $P_i$, after populating a subject that corresponds to the relation $r_i$ and a mask token. We expect the model to make the same prediction for the masked token for all patterns.

\paragraph{Consistency Loss Function}
As we evaluate the model using acc@1, the straight-forward consistency loss would require these predictions to be identical:
\begin{gather*} 
\min_{\theta} \mbox{sim}(\arg\max_i f_\theta(P_n)[i], \arg\max_j f_\theta(P_m)[j])%
\end{gather*}
where $f_\theta(P_n)$ is the output of an encoding function (e.g., BERT) parameterized by $\theta$ (a vector) over input $P_n$, and $f_\theta(P_n)[i]$ is the score of the $i$th vocabulary item of the model.

However, this objective contains a comparison between the output of two argmax operations, making it discrete and discontinuous, and hard to optimize in a gradient-based framework. We instead relax the objective, and require that the predicted \emph{distributions} $Q_n = \mbox{softmax}(f_\theta(P_n))$, rather than the top-1 prediction, be identical to each other. %
We use two-sided KL Divergence to measure similarity between distributions: $D_{KL}(Q_n^{r_i}||Q_m^{r_i}) + D_{KL}(Q_m^{r_i}||Q_n^{r_i})$
where $Q_n^{r_i}$ is the predicted distribution for pattern $P_n$ of relation $r_i$.

As most of the vocabulary is not relevant for the
predictions, we filter it down to the $k$ tokens from the candidate set of each
relation (\S\ref{sec:framework}). We want to
maintain the original capabilities of the
model -- focusing on the candidate set helps to achieve this goal since most of the vocabulary is not affected by our new
loss.

To encourage a more general solution, we make use of all the paraphrases together, and enforce all predictions to be as close as possible.
Thus, the consistency loss for all pattern pairs for a particular relation $r^i$ is:
\[
\mathcal{L}_{c} = \sum_{n=1}^k \sum_{m=n+1}^k D_{KL}(Q^{r_i}_n||Q^{r_i}_m) + D_{KL}(Q^{r_i}_m||Q^{r_i}_n)
\]

\paragraph{MLM Loss}
Since the consistency loss is different from the
Cross-Entropy loss the PLM is trained on, we find it
important to continue the MLM loss on text data, similar to previous work \cite{geva2020injecting}.

We consider two alternatives for continuing the pretraining objective: (1) MLM on Wikipedia and (2) MLM on the patterns of the relations used for the consistency loss. We found that the latter works better. We denote this loss by $\mathcal{L}_{MLM}$.

\paragraph{Consistency Guided MLM Continual Training}

Combining our novel consistency loss with
the regular MLM loss, we continue the PLM training by
combining the two losses. The combination of the two losses
is determined by a hyperparameter $\lambda$, resulting in
the following final loss function:
\[
\mathcal{L} = \lambda \mathcal{L}_c + \mathcal{L}_{MLM}
\]
This loss is computed per relation, for one KB tuple. We have many of these instances, which we require to behave similarly. Therefore, we batch together $l=8$ tuples from the same relation and apply the consistency loss function to all of them.

\subsection{Setup}

Since we evaluate our method on unseen relations, we also
split train and test by relation type (e.g., location-based relations, which are very common
in T-REx).  Moreover, our method is aimed to be simple,
effective, and to require only minimal supervision. Therefore,
we opt to use only three relations:
\textit{original-language}, \textit{named-after}, and
\textit{original-network}; these were chosen randomly, out of
the non-location related relations.\footnote{Many relations are location-based -- not training on them prevents train-test leakage.} 
For validation, we randomly pick three
relations of the remaining relations and use the remaining
25 for testing.

We perform minimal tuning of the parameters ($\lambda \in {0.1, 0.5, 1}$) to pick the best model, train for three epochs, and select the best model based on  \textit{Consistent-Acc} on the validation set.
For efficiency reasons, we use the base version of BERT.

\subsection{Improved Consistency Results}

\begin{table}[t]
    \centering
\resizebox{1\columnwidth}{!}{%
\begin{tabular}{lrrr}
\toprule
Model &        Accuracy & Consistency & Consistent-Acc \\
\midrule

majority   &  24.4+-22.5 &  100.0+-0.0 &  24.4+-22.5 \\
\midrule
BERT-base  &  45.6+-27.6 &  58.2+-23.9 &  27.3+-24.8 \\
BERT-ft    &  \textbf{\textul{47.4}}+-27.3 &  \textbf{64.0}+-22.9 &  \textbf{\textul{33.2}}+-27.0 \\
\quad -consistency &  46.9+-27.6 &  60.9+-22.6 &  30.9+-26.3 \\
\quad -typed     &  46.5+-27.1 &  62.0+-21.2 &  31.1+-25.2 \\
\quad -MLM       &  16.9+-21.1 &  \textul{80.8}+-27.1 &   9.1+-11.5 \\

\bottomrule
\end{tabular}

}

    \caption{Knowledge and consistency results for the baseline, BERT base, and our model. 
    The results are averaged over the 25 test
    relations.
Underlined: best performance
    overall, including  ablations.
    Bold: best performance for  BERT-ft and the two
    baselines (BERT-base, majority).}
    \label{tab:consistency-ft}
    
    \vspace{-0.2in}
\end{table}

The results are presented in Table
\ref{tab:consistency-ft}. We report aggregated results
for the 25 relations in the test.
We again
report macro average (mean over relations) and
standard deviation.  We report the results of the majority
baseline (first row),   BERT-base  (second row)
and our new model (BERT-ft, third row).  First, we note
that our model significantly improves consistency: 64.0\% (compared with 58.2\% for BERT-base,
an increase
of 5.8 points).  \textit{Accuracy} also improves compared to BERT-base, from 45.6\% to 47.4\%. Finally, and most
importantly, we see an increase of 5.9 points in
\textit{Consistent-Acc}, which is achieved due to the improved
consistency of the model.  Notably, these improvements
arise from training on merely three relations, meaning that
the model improved its consistency ability and generalized
to new relations.  We measure the statistical
significance of our method compared to the BERT baseline,
using McNemar's test (following
\citet{dror2018hitchhiker,dror2020statistical}) and find all
results to be significant ($p \ll 0.01$).

We also perform an ablation study to quantify the utility of
the different components. First, we report on the finetuned
model without the consistency loss
(-consistency). Interestingly, it does improve over the
baseline (BERT-base), but it lags behind our finetuned model.
Second, applying our loss on the candidate set rather than
on the entire vocabulary is beneficial (-typed). Finally, by
not performing the MLM training on the generated patterns
(-MLM), the consistency results improve significantly
(80.8\%); however, this also hurts  \textit{Accuracy} and \textit{Consistent-Acc}.
MLM training seems to serve as a regularizer
that prevents catastrophic forgetting.

Our ultimate goal is to improve consistency in PLMs for better performance on downstream tasks. Therefore, we also experiment with finetuning on SQuAD \cite{squad}, and evaluating on paraphrased questions from SQuAD \cite{squad-paraphrase} using our consistency model. However, the results perform on par with the baseline model, both on SQuAD and the paraphrase questions. More research is required to show that consistent PLMs can also benefit downstream tasks.

\section{Discussion}
\label{sec:discussion}

\paragraph{Consistency for Downstream Tasks}

The rise of PLMs has improved many tasks but has also brought a lot of expectations. The standard usage of these models is pretraining on a large corpus of unstructured text and then finetuning on a task of interest. The first step is thought of as providing a good language-understanding component, whereas the second step is used to teach the format and the nuances of a downstream task.

As discussed earlier, consistency is a crucial component of many NLP systems \cite{du2019consistent,consistent-qa,denis2009global,kryscinski2020evaluating} and obtaining this skill from a pretrained model would be extremely beneficial and has the potential to make specialized consistency solutions in downstream tasks redundant.
Indeed, there is an ongoing discussion about the ability to acquire ``meaning'' from raw text signal alone \cite{bender2020climbing}.
Our new benchmark makes it possible to track the progress of consistency in pretrained models.

\paragraph{Broader Sense of Consistency}
In this work we focus on one type of consistency, that is,
consistency in the face of paraphrasing; however, consistency is
a broader concept.  For instance, previous work has studied
the effect of negation on factual statements, which can also
be seen as consistency
\cite{Ettinger_2020,kassner-schutze-2020-negated}. 
A consistent model is expected to return  different answers
to the prompts: ``\textit{Birds} can \textit{[MASK]}'' and
``\textit{Birds} cannot \textit{[MASK]}''. The inability to
do so, as was shown in these works, also shows the lack of
model consistency.

\paragraph{Usage of PLMs as KBs}
Our work follows the setup of \citet{lama,alpaqa}, where PLMs are being tested as KBs. While it is an interesting setup for probing models for knowledge and consistency, it lacks important properties of standard KBs: (1) the ability to return more than a single answer and (2) the ability to return no answer.
Although some heuristics can be used for allowing a PLM to do so, e.g., using a threshold on the probabilities, it is not the way that the model was trained, and thus may not be optimal.
Newer approaches that propose to use PLMs as a starting point to more complex systems have promising results and address these problems \cite{thorne2020neural}.

In another approach,  \citet{autoprompt} suggest using \textsc{AutoPrompt} to automatically generate prompts, or patterns, instead of creating them manually. This approach is superior to manual patterns \cite{lama}, or aggregation of patterns that were collected automatically \cite{alpaqa}.

\paragraph{Brittleness of Neural Models}
Our work also relates to the problem of the brittleness of neural networks. One example of this brittleness is the vulnerability to adversarial attacks \cite{adversarial_attacks,jia2017adversarial}.
The other problem, closer to the problem we explore in this work, is the poor generalization to paraphrases.
For example, \citet{squad-paraphrase} created a paraphrase version for a subset of SQuAD \cite{squad}, and showed that model performance drops significantly. 
\citet{ribeiro2018semantically} proposed another method for
creating paraphrases and performed a similar analysis for
visual question answering and sentiment analysis. Recently,
\citet{ribeiro-etal-2020-beyond} proposed
\textsc{CheckList}, a system that tests a model's vulnerability to several linguistic perturbations.

\resource{} enables us to study the brittleness of PLMs, and
separate  facts that are robustly encoded in the model from
mere `guesses', which may arise from some heuristic or
spurious correlations with certain patterns
\cite{poerner2020bert}. We showed that PLMs are susceptible
to small perturbations, and thus, finetuning on a
downstream task -- given that training datasets are typically not
large and do not contain equivalent examples -- is not
likely to perform better.

\paragraph{Can we Expect from LMs to be Consistent?}

The typical training procedure of an LM does not encourage consistency. The standard training solely tries to minimize the log-likelihood of an unseen token, and this objective is not always aligned with consistency of knowledge. Consider for example the case of wikipedia texts, as opposed to reddit; their texts and styles  may be very different and they may even describe contradictory facts. An LM can exploit the styles of each text to best fit the probabilities given to an unseen word, even if the resulting generations contradict each other.

Since the pretraining-finetuning procedure is the dominating one in our field currently, a great amount of the language capabilities that were learned during pre-training also propagates to the fine-tuned models. As such, we believe it is important to measure and improve consistency already in the pretrained models.

\paragraph{Reasons Behind the (In)Consistency}

Since LMs are not expected to be consistent, what are the reasons behind their predictions, when being consistent, or inconsistent?

In this work, we presented the predictions of multiple queries, and the representation space of one of the inspected models. However, this does not point to the origins of such behavior.
In future work, we aim to inspect this question more closely.

\section{Conclusion}
\label{sec:conclusions}

In this work, we study the consistency of PLMs with regard to their ability to extract knowledge.
We build a high-quality resource named \resource{} that contains 328 high-quality patterns for 38 relations.
Using \resource{}, we measure consistency in multiple PLMs,
including BERT, RoBERTa, and ALBERT, and show that although
the latter two are superior to BERT in other tasks, they
fall short in terms of consistency. However, the consistency
of these models is generally low.
We release \resource{} along with data tuples from T-REx as
a new benchmark to track knowledge consistency of NLP models.
Finally, we propose a new simple method to improve model consistency, by continuing the pretraining with a novel loss. We show this method to be effective and to improve both the consistency of models as well as their ability to extract the correct facts.

\section*{Acknowledgements}
We would like to thank Tomer Wolfson, Ido Dagan, Amit Moryossef and Victoria Basmov for their helpful comments and discussions, and Alon Jacovi, Ori Shapira, Arie Cattan, Elron Bandel, Philipp Dufter, Masoud Jalili Sabet,  Marina Speranskaya, Antonis Maronikolakis, Aakanksha Naik, Aishwarya Ravichander, Aditya Potukuchi for the help with the annotations.
We also thank the anonymous reviewers and the action editor, George Foster, for their valuable suggestions.

Yanai Elazar is grateful to be supported by the PBC fellowship for outstanding PhD candidates in Data Science and the Google PhD fellowship.
This project has received funding from the Europoean Research Council (ERC) under the Europoean Union's Horizon 2020 research and innovation programme, grant agreement No. 802774 (iEXTRACT).
This work has been funded by the European Research Council (\#740516) and by the German Federal Ministry of Education
and Research (BMBF) under Grant No. 01IS18036A. The authors of this
work take full responsibility for its content. This research was also supported in part by grants from the National Science Foundation Secure and Trustworthy Computing program (CNS-1330596, CNS15-13957, CNS-1801316, CNS-1914486) and a DARPA Brandeis grant (FA8750-15-2-0277). The views and conclusions contained herein are those of the authors and should not be interpreted as necessarily representing the official policies or endorsements, either expressed or implied, of the NSF, DARPA, or the US Government.

\bibliography{custom}
\bibliographystyle{acl_natbib}

\appendix

\section{Appendix}
\label{sec:appendix}

We heavily rely on Hugging Face's Transformers library \cite{wolf-etal-2020-transformers} for all experiments involving the PLMs.
We used Weights \& Biases for tracking and logging the experiments \cite{wandb}.
Finally, we used sklearn \cite{scikit-learn} for other ML-related experiments.

\section{Paraphrases Analysis}
\label{sec:paraphrase_analysis}

We provide a characterization of the paraphrase types included in our dataset. 

We analyze the type of paraphrases in \resource{}. We sample 100 paraphrase pairs from the agreement evaluation that were labeled as paraphrases and annotate the paraphrase type. Notice that the paraphrases can be complex, as such, multiple transformations can be annotated for each pair.
We mainly use a subset of paraphrase types categorized by \citet{what_is_paraphrase}, but also define new types which were not covered by that work.
We begin by briefly defining the types of paraphrases found in \resource{} from \citet{what_is_paraphrase} (more thorough definitions can be found in their paper), and then define the new types we observed.

\begin{table*}[t!]
    \centering
\resizebox{1\textwidth}{!}{%
\begin{tabular}{lllllllll}
\toprule
Paraphrase Type & Pattern \#1 & Pattern \#2 & Relation & N. \\
\midrule

Synonym substitution & [X] died in [Y]. & [X] expired at [Y]. & place of death & 41 \\

Function words variations & [X] is [Y] citizen. & [X], who is a citizen of [Y]. & country of citizenship & 16 \\

Converse substitution & [X] maintains diplomatic relations with [Y]. & [Y] maintains diplomatic relations with [X]. & diplomatic relation & 10 \\

Change of tense & [X] is developed by [Y]. & [X] was developed by [Y]. & developer & 10 \\ 

Change of voice & [X] is owned by [Y]. & [Y] owns [X]. & owned by & 7 \\

Verb/Noun conversion & The headquarter of [X] is in [Y]. & [X] is headquartered in [Y]. & headquarters location & 7 \\

External knowledge & [X] is represented by music label [Y]. & [X], that is represented by [Y]. & record label & 3 \\

Noun/Adjective conversion & The official language of [X] is [Y]. & The official language of [X] is the [Y] language. & official language & 2 \\

Change of aspect & [X] plays in [Y] position. & playing as an [X], [Y] & position played on team & 1 \\

\midrule

Irrelevant addition & [X] shares border with [Y]. & [X] shares a common border with [Y]. & shares border with & 11 \\

Topicalization transformation & [X] plays in [Y] position. & playing as a [Y], [X] & position played on team & 8 \\

Apposition transformation & [X] is the capital of [Y]. & [Y]'s capital, [X]. & capital of & 4 \\ 

Other syntactic movements & [X] and [Y] are twin cities. & [X] is a twin city of [Y]. & twinned administrative body & 10 \\

\bottomrule
\end{tabular}

}
\caption{Different types of paraphrases in \resource{}. We report examples from each paraphrase type, along with the type of relation, and the number of examples from the specific transformation from a random subset of 100 pairs. Each pair can be classified into more than a single transformation (we report one for brevity), thus the sum of transformation is more than 100.}
\label{tab:paraphrases_analysis}
    
\end{table*}

\begin{enumerate}
    \item Synonym substitution: Replacing a word/phrase by a synonymous word/phrase, in the appropriate context.
    \item Function word variations: Changing the function words in a sentence/phrase without affecting its semantics, in the appropriate context.
    \item Converse substitution: Replacing a word/phrase with its converse and inverting the relationship between the constituents of a sentence/phrase, in the appropriate context, presenting the situation from the converse perspective.
    \item Change of tense: Changing the tense of a verb, in the appropriate context.
    \item Change of voice: Changing a verb from its active to passive form and vice versa results in a paraphrase of the original sentence/phrase.
    \item Verb/Noun conversion: Replacing a verb by its corresponding nominalized noun form and vice versa, in the appropriate context.
    \item External knowledge: Replacing a word/phrase by another word/phrase based on extra-linguistic (world) knowledge, in the appropriate context.
    \item Noun/Adjective conversion: Replacing a verb by its corresponding adjective form and vice versa, in the appropriate context.
    \item Change of aspect: Changing the aspect of a verb, in the appropriate context.
    
\end{enumerate}

We also define several other types of paraphrases not covered in \citet{what_is_paraphrase} (potentially because they did not occur in the corpora they have inspected). 

\begin{enumerate}[a.]
\item Irrelevant addition: addition or removal of a word or phrase, that does not affect the meaning of the sentence (as far as the relation of interest is concerned), and can be inferred from the context independently.
\item Topicalization transformation: a transformation from or to a topicalization construction. Topicalization is a construction in which a clause is moved to the beginning of its enclosing clause.
\item Apposition transformation: a transformation from or to an apposition construction. In an apposition construction, two noun phrases where one identifies the other are placed one next to each other.
\item Other syntactic movements: includes other types of syntactic transformations that are not part of the other categories. This includes cases such as moving an element from a coordinate construction to the subject position as in the last example in Table \ref{tab:paraphrases_analysis}. Another type of transformation is in the following paraphrase: ``[X] plays in [Y] position.'' and ``[X] plays in the position of [Y].'' where a compound noun-phrase is replaced with a prepositional phrase.
\end{enumerate}

We report the percentage of each type, along with examples of paraphrases in Table \ref{tab:paraphrases_analysis}.
The most common paraphrase is the `synonym substitution', following `function words variations' which occurred 41 and 16 times, respectively. The least common paraphrase is `change of aspect', which occurred only once in the sample.

The full \resource{} resource can be found at: \url{https://github.com/yanaiela/pararel/tree/main/data/pattern_data/graphs_json}.

\end{document}